\RequirePackage{fix-cm}
\documentclass[twocolumn]{svjour3}          
\smartqed  
\usepackage{url}
\usepackage{graphicx}
\usepackage{amssymb}
\usepackage{amsmath}
\usepackage{booktabs}
\usepackage{multicol}
\usepackage{multirow}
\usepackage{mathptmx}   
\usepackage{caption}
\usepackage{float}
\usepackage{stfloats}
\usepackage{xurl}
\usepackage{hyperref}
\usepackage{latexsym}
\journalname{The Visual Computer}
\begin{document}

\title{JVLGS: Joint Vision–Language Gas Leak Segmentation}

\author{Xinlong Zhao\textsuperscript{1} \and 
       Qixiang Pang\textsuperscript{2} \and Shan Du\textsuperscript{1} }

\institute{
     Shan Du * \\
     shan.du@ubc.ca\\[1ex]
     Xinlong Zhao \\
     zhaoxinlong1@gmail.com \\[1ex]
     Qixiang Pang \\                
     qpang@ucmo.edu \\[1ex]
    \textsuperscript{1}Irving K. Barber Faculty of Science, University of British Columbia Okanagan, Kelowna, BC V1V 1V7, Canada \\ [1ex]
    \textsuperscript{2}Department of Computer Science and Cybersecurity, University of Central Missouri, Warrensburg, MO 64093, USA
}

\date{Received: date / Accepted: date}

\maketitle

\begin{abstract}
Gas leaks pose severe risks to human health and industrial safety. However, accurate and timely monitoring of gas leaks remains a major challenge. Existing vision-based methods using infrared (IR) imagery are limited by the inherently blurry and non-rigid nature of leak plumes, which reduces detection reliability and precision. To overcome these limitations, this paper proposes a Joint Vision–Language Gas leak Segmentation (JVLGS) framework that integrates the complementary strengths of visual and textual modalities to enhance gas leak segmentation. Recognizing that gas leaks are sporadic and many video frames contain no leakage, JVLGS incorporates an adaptive postprocessing module to effectively suppress false positives caused by noise and non-target objects—a common limitation of existing approaches. Extensive experiments across diverse industrial scenarios de-monstrate that JVLGS significantly outperforms state-of-the-art gas leak segmentation methods. Furthermore, it achieves consistently strong performance under both supervised and few-shot learning settings, whereas competing methods typically perform well in only one setting or underperform in both. We publish our code at: \url{https://github.com/GeekEagle/JVLGS}.

\end{abstract}

\section{Introduction}
Industrial gas leaks constitute critical emergencies that can lead to severe accidents. Methane contributes significantly to the greenhouse effect, while others like acetaldehyde are highly toxic \cite{VOCleakdetection}. Detecting these leaks is especially challenging because most industrial gases are colorless and odorless, making them imperceptible to the human senses. Traditional detection methods rely on electrochemical or photoionization sensors to monitor pipeline conditions \cite{yuan2023leak}. These approaches are hazardous because they require inspectors to be close to suspected leak points. To mitigate these hazards, vision-based detection methods have gained increasing attention due to their non-contact nature and potential for automation \cite{ying2020optical}. Such methods reduce labor costs while eliminating direct exposure risks for inspectors \cite{ravikumar2017opticaleff}. Among them, video surveillance integrated with infrared (IR) imaging has shown strong potential for detecting gas leaks.

Nevertheless, video surveillance systems still face significant challenges. Gas leaks are difficult to detect, even with infrared cameras, because they blend seamlessly into the background. To address this issue, Lu et al. \cite{lu2021effective} proposed a Gaussian-based background modeling method for adaptive gas leak segmentation. However, its effect degrades significantly in the presence of irregular gas shapes and dynamic backgrounds. Similarly, optical flow-based methods \cite{yang2021mg} are limited by inconsistent gas shapes and are susceptible to noise interference.

Advances in deep learning have led to significant progress in camouflaged object detection (COD) \cite{xiao2024codsurvey}. For example, SINet-V2 \cite{fan2021sinetv2} outlined the rough contours of concealed objects, and GasSeg \cite{yu2024gasseg} combined convolutional neural networks (CNNs) with Vision Transformers (ViT) to develop a lightweight model for gas leak segmentation. However, they struggle to capture motion cues from faint and ambiguous targets. Long-term motion-based techniques \cite{xmempp,cheng2024cutie} face difficulties due to the highly variable shapes of gas leaks and the lack of consistent features, while short-term methods, such as RCRNet \cite{yan2019rcrnet}, leverage optical flow but remain susceptible to noise. Implicit motion analysis methods, like SLT-Net \cite{cheng2022sltnet}, show potential but struggle to distinguish gas leaks from texturally similar objects, such as clouds. FGSTP \cite{zhao_fgstp} introduced fine-grained perception to refine motion information; however, the extracted features are not always reliably correlated with actual targets.

To enhance target perception under low-contrast infrared conditions, recent infrared-visible image fusion studies provide useful insights into complementary feature modeling and local feature enhancement. For instance, IVOMFuse \cite{zhang2023ivomfuse} introduces infrared-to-visible object mapping for object-aware fusion, FusionNGFPE \cite{zhang2025fusionngfpe} adopts non-global enhancement to strengthen local features before fusion, and synthetic infrared visible data \cite{gu2023infrareddata} generated from game engines has been used to improve fusion models when real paired data are limited. These studies suggest that image fusion approaches can improve the representation of weak targets in challenging infrared scenes. Nevertheless, most fusion methods aim to generate visually enhanced images from paired infrared-visible inputs, whereas gas leak segmentation requires frame-wise dense mask prediction for blurry, deformable, and temporally unstable gas plumes. Therefore, directly adopting image fusion strategies is insufficient for accurate gas leak segmentation.

Beyond visual enhancement, another promising direction is to introduce semantic guidance into temporal-spatial representation learning. Vision-Language Models (VLMs) \cite{vlmsurvey} have demonstrated strong potential in aligning visual features with textual concepts. For instance, CLIP \cite{clip2021} uses a ViT as both image and text encoder, but is too slow for real-time segmentation. VisualBERT \cite{li2019visualbert} enables end-to-end vision-language correlation learning but lacks spatial precision. DALL-E \cite{dalle2} combines visual information with large language models (LLMs) for creative generation but is unsuitable for precise segmentation. OWLv2 \cite{owlv2} employs a ViT to process both image and text inputs via self-training and shows potential for open-vocabulary detection. Nevertheless, few existing methods effectively incorporate text-guided segmentation for blurry, non-rigid objects such as gas leaks.

In this paper, we propose Joint Vision-Language Gas leak Segmentation (JVLGS), a novel framework designed to enhance model generalizability across diverse gas leak scenarios. Our method integrates text prompts between the vision encoder and motion extractor to enrich feature representations. Following vision–language integration, a temporal-spatial module is designed to analyzes motion and spatial cues of gas leaks using these enhanced features. The decoder then integrates multi-scale information through upsampling to generate the final masks. Many prior methods \cite{pang2024zoomnext,cheng2022sltnet} are vulnerable to noise and minor movements in non-leak frames, often producing false positives when no target is present. To address this, JVLGS incorporates a post-processing step that eliminates false positives, ensuring completely blank masks for non-leak cases. We evaluate our model on two benchmark datasets under both supervised and few-shot learning settings to simulate realistic video surveillance scenarios.

The main contributions of this work are summarized as follows:
\begin{itemize}
  \item We propose JVLGS, a framework that integrates manually crafted text prompts with a video clip to enhance feature extraction. These prompts enable accurate localization and detection of gas leaks.
  \item We design a lightweight vision-language fusion approach, enabling more effective integration of visual features with textual representations.
  \item JVLGS achieves significant performance gains in both supervised and few-shot learning, effectively simulating real-world leak detection scenarios. 
  \item Among all benchmark models, JVLGS delivers a favorable trade-off between segmentation accuracy and computational efficiency, making it suitable for practical industrial monitoring applications.
\end{itemize}

\section{Related Work}
\subsection{Gas Leak Detection and Segmentation}
Traditional gas leak detection and segmentation methods rely on manual sampling and laboratory analysis based on chemical or stoichiometric reactions. Although these approaches are generally accurate, they suffer from slow response times, high labor costs, and safety risks associated with manual gas sampling \cite{tradlimits}.

Recent advancements in infrared (IR) thermography have made vision-based detection a promising alternative \cite{yuan2023leak}. Vision-based methods offer convenient operation, wide detection coverage, real-time feedback, and intuitive visualization of gas leaks. However, automatically segmenting invisible gas remains challenging. Existing vision-based methods can be broadly classified into image-based and video-based approaches. Image-based methods \cite{xu2019flame,ye2015dynamic} depend on local visual features, such as shape and texture. However, their performance degrades when dealing with blurry or low-contrast gas regions in infrared imagery. In contrast, video-based methods \cite{wang2022videogasnet,kopbayev2022gas} exploit the motion characteristics of gas in video sequences, which typically improves detection accuracy. Nevertheless, these approaches often incur high computational costs and are unsuitable for real-time industrial applications. To overcome these limitations, we propose a video-based framework that effectively integrates visual and textual information to enhance gas leak segmentation. The proposed model achieves state-of-the-art performance while maintaining practical efficiency for industrial deployment.

\subsection{Camouflaged Object Detection}
Camouflaged Object Detection (COD) aims to identify objects that blend seamlessly into their surroundings, often exhibiting low contrast and ambiguous boundaries. Traditional COD methods \cite{tracod_1} rely on hand-crafted features such as texture, color, or edge cues. However, these approaches perform poorly in complex environments or low-resolution imagery due to their limited representational capacity. With the advent of deep learning, bioinspired approaches have been introduced to enhance detection efficiency and robustness \cite{yan2021mirrornet}. Although these methods achieve real-time inference, they remain less effective for detecting small or blurry objects.

To address these limitations, motion-based COD approa-ches further exploit temporal information by incorporating cross-frame cues and motion coherence modules to improve segmentation precision \cite{he2023strategic,lamdouar2020betrayed}. To overcome the limitations of two-stage motion-based pipelines, recent end-to-end models perform object localization and segmentation within a unified framework, thereby suppressing distractions from background motion \cite{hui2024endow,zhang2025emip}. These models leverage temporal continuity to ensure inter-frame consistency of detected objects.

Furthermore, recent light-field occlusion removal studies provide another relevant perspective on recovering visually degraded or partially hidden structures \cite{LFORreview-ieee,advanceslfor}. Existing methods such as DAORNet\cite{senussi2026daornet}, U2-LFOR \cite{u2lfor}, LF-PyrNet \cite{senussi2025lr-pyrnet}, and TriORU2-Net++ \cite{senussi2026trioru2} exploit multi-scale receptive fields, feature pyramid aggregation, dense attention, or U-shaped hierarchical refinement to reconstruct occluded regions from light-field images. These studies demonstrate that ambiguous visual structures often require multi-level contextual reasoning \cite{cspdarknet53-orlf} and refinement rather than relying only on local appearance cues \cite{senussi2025spectral}. However, light-field occlusion removal is fundamentally different from infrared gas leak segmentation: it usually assumes multi-view light-field inputs and aims to reconstruct occluded content, whereas our task uses monocular infrared video to produce frame-wise binary masks for non-rigid, semi-transparent gas plumes.

Inspired by these advancements, this paper proposes an end-to-end framework specifically designed for gas leak monitoring in complex and unstable industrial environments. In contrast to existing COD and light-field occlusion removal methods, JVLGS combines text-guided feature enhancement with temporal-spatial motion reasoning to segment gas leaks in infrared videos. The proposed approach effectively captures subtle, motion-dependent leak patterns that closely resemble camouflaged objects in dynamic infrared scenes.

\subsection{Vision-Language Model}
Vision–Language Models (VLMs) map both visual and textual modalities into a shared embedding space, thereby enhancing the visual features \cite{vlmsurvey}. CLIP \cite{clip2021} is a representative dual-encoder model that learns image-text alignment through contrastive pretraining. Based on CLIP, CLIPSeg \cite{luddecke2022clipseg} introduces a decoder to generate masks from text or image prompts, and DenseCLIP \cite{rao2021denseclip} converts image-text matching into pixel-text matching for dense prediction. Although these methods demonstrate the benefit of textual semantics, they mainly rely on image-level or pixel-level semantic alignment in single images. For infrared gas leak segmentation, such alignment is limited because gas plumes are blurry, non-rigid, semi-transparent, and lack stable object boundaries.

Open-vocabulary detection methods, such as OWL-ViT \cite{minderer2022owlvit} and OWLv2 \cite{owlv2}, match textual queries with visual regions for object localization. However, they are primarily designed for box-level detection rather than dense mask prediction. This makes them less suitable for gas leaks, whose shapes are irregular and continuously changing. Therefore, we do not directly use OWLv2 as a detector, but repurpose its text module as a semantic prompt encoder.

Recent prompt-guided and referring segmentation methods further study cross-modal interaction between language and visual features. For example, Prompt-RIS \cite{shang2024promptRIS} introduces cross-modal prompting for referring image segmentation, LQMFormer \cite{shah2024lqmformer} models language-aware query masks, and Mask Grounding \cite{chng2024maskground} investigates mask-level language-image alignment. These methods usually assume that the text directly refers to an observable object or region. This assumption is weakened in gas leak segmentation, where “gas leak” is not a visually explicit category in pretrained VLMs and the target lacks stable texture, contour, and identity.

Different from the above methods, JVLGS uses language as a semantic prior rather than as a direct detection or mask query. This design enhances weak leak-related responses while preserving the spatial structure required for dense video segmentation.

\begin{figure*}[t]
    \centering
    \includegraphics[width=\textwidth]{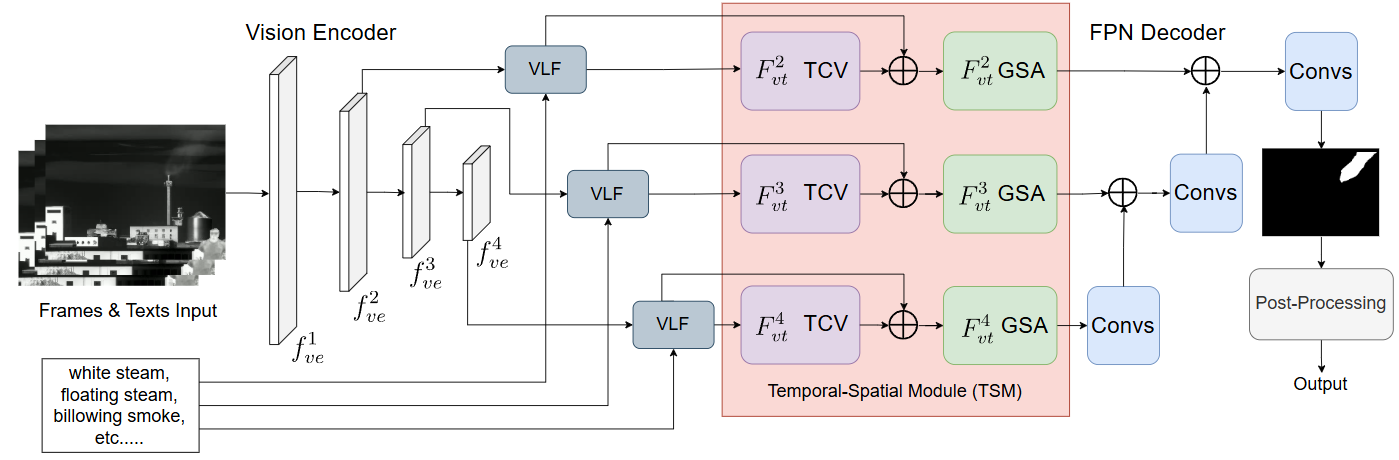}
    \caption{Framework of Joint Vision-Language Gas leak Segmentation (JVLGS). The inputs are video clips and text prompts of the target object. Visual and textual features are fused in the VLF module, followed by the Temporal-Spatial Module (TSM) and decoder. The post-processing removes false positives for final masks.}
    \label{framework}
\end{figure*}

\section{Method}
\subsection{Model Architecture}
Fig. 2 illustrates the overall architecture of the proposed Joint Vision–Language Gas leak Segmentation (JVLGS) framework. The model comprises the following components: (1) Vision encoders based on Pyramid Vision Transformer-v2 (PVTv2) \cite{wang2022pvtv2}; (2) A Vision-Language Fusion (VLF) module designed to integrate visual features with text representations effectively; (3) A Temporal-Spatial Module (TSM) proposed to capture fine-grained motion and spatial information related to target objects; (4) An enhanced decoder built upon the Feature Pyramid Network (FPN) \cite{fpn_original}, optimized for multi-scale feature decoding; (5) A post-processing step to remove false positives. Overall, JVLGS takes video clips and text prompts as inputs and generates a binary mask for each frame, indicating the presence or absence of gas leakage.

Numerous studies have demonstrated that text prompts can enrich semantic representations, enabling models to better distinguish target objects in complex visual scenes \cite{action,dshmp}. Our proposed JVLGS integrates textual representations into visual features, enabling the Temporal-Spatial Module (TSM) to localize targets more accurately through text-guided encoding. Furthermore, we fine-tune the pretrained text model of OWLv2 \cite{owlv2} as our text module.

\subsection{Vision-Language Fusion}
\begin{figure}
    \centering
    \includegraphics[width=0.9\linewidth]{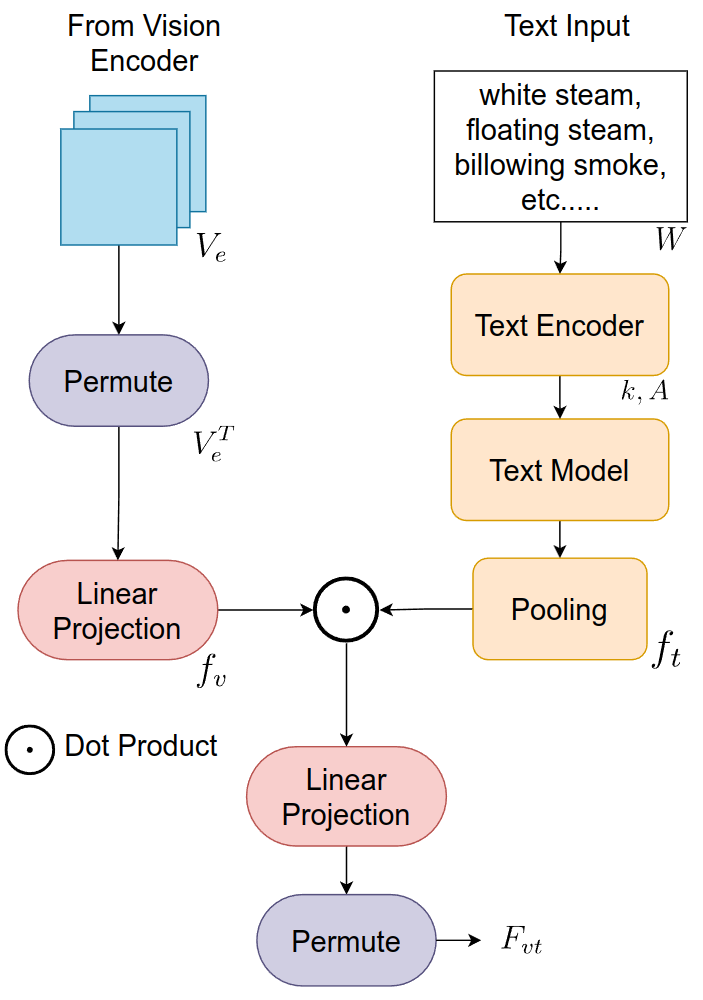}
    \caption{Architecture of the Vision-Language Fusion (VLF) module. The vision encoder extracts visual features, while text prompts are generated from keywords or synonyms of the target objects.}
    \label{language prompt}
\end{figure}

The input video clips are first processed and divided by the vision encoder, PVTv2 \cite{wang2022pvtv2} into three scale features $f_{ve}^i \in R^{C_v\times H/2^{i+1} \times W/2^{i+1}}, i\in \{2,3,4\}$, where $C_v, H, W$ are channels, height, and width of a frame. Fig. \ref{language prompt} shows the details of the Vision-Language Fusion (VLF) module. The input of VLF, $V_e$, consists of $f_{ve}^i$. The size of $V_e$ is $T\times C_v\times H/2^{i+1}\times W/2^{i+1}$, where $T$ is the length of the video clip. The VLF firstly permute $V_e$ to $V_e^T \in R^{T\times H/2^{i+1} \times W/2^{i+1} \times C_v}$. Then $V_e^T$ is linearly transformed to the vision feature $f_{v}$.
 
The text encoder in JVLGS is adapted from the OWLv2 text module, which employs a Vision Transformer (ViT) backbone and was originally developed for zero-shot object detection \cite{owlv2}. Although OWLv2 outputs bounding boxes, in our framework, its textual embeddings are repurposed to guide visual attention toward target regions, thereby enabling the generation of more precise binary segmentation masks. The VLF module integrates visual cues with textual guidance, where vision features support target-object localization and text representations enrich these features. Considering the high variability of industrial video content and the linguistic limitations of pretrained language models, multiple descriptive prompt words are utilized to capture diverse gas leak patterns. 

Since gas leaks are often invisible to the human eye and standard RGB cameras, many pretrained vision-language models do not possess a direct and reliable visual-semantic representation of the concept “gas leak”. However, in infrared videos, gas leakage regions usually exhibit visual patterns similar to smoke- or steam-like phenomena, such as drifting, flowing, billowing, or semi-transparent structures. Therefore, instead of directly using “gas leak” as the only textual prompt, we empirically select several text prompts that describe gas-like visual characteristics, including smoke- and steam-related expressions. These prompts provide the language module with more accessible semantic cues. Based on our experimental comparisons, we finally choose the four most relevant prompts that consistently provide useful guidance for gas leak segmentation. Further analyses of the selected prompts are provided in Section $4.4$. We denote the text input as $W \in R^{P_l\times d}$, where $P_l$ is the number of the prompt sequence and $d$ is the dimension of each prompt sequence. Then the prompted words are encoded by the text encoder as follows: 
\begin{align}
 k,A = TextEncoder(W)
\end{align}
where $k$ is the key of the prompt sequences, and $A$ is an attention mask, indicating which tokens are correlated inputs for the text model. The $k$ and $A$ are then fed into the text encoder to obtain the textual representations as follows:
\begin{align}
f_{t} = TP(OwlText(k,A))
\end{align}
where $OwlText(k, A)$ represents the extracted text representations $f_t \in R^{P_l\times C_t}$, and $C_t$ is the channel size of text representation. $TP(\cdot)$ denotes token-level pooling that selects a token from the hidden state of each sequence to refine $f_t$. Next, the text representation fuses with the $f_v$. The fusion operation in each scale is as follows:
\begin{align}
F_{vt} = (\phi(f_v \cdot f_t))^T
\end{align}
where $\phi(\cdot)$ is the linear projection that converts the channel number of representative text $C_{t}$, to the channel number of vision features, $C_{v}$.

\begin{figure*}
    \centering
    \includegraphics[width=1.0\linewidth]{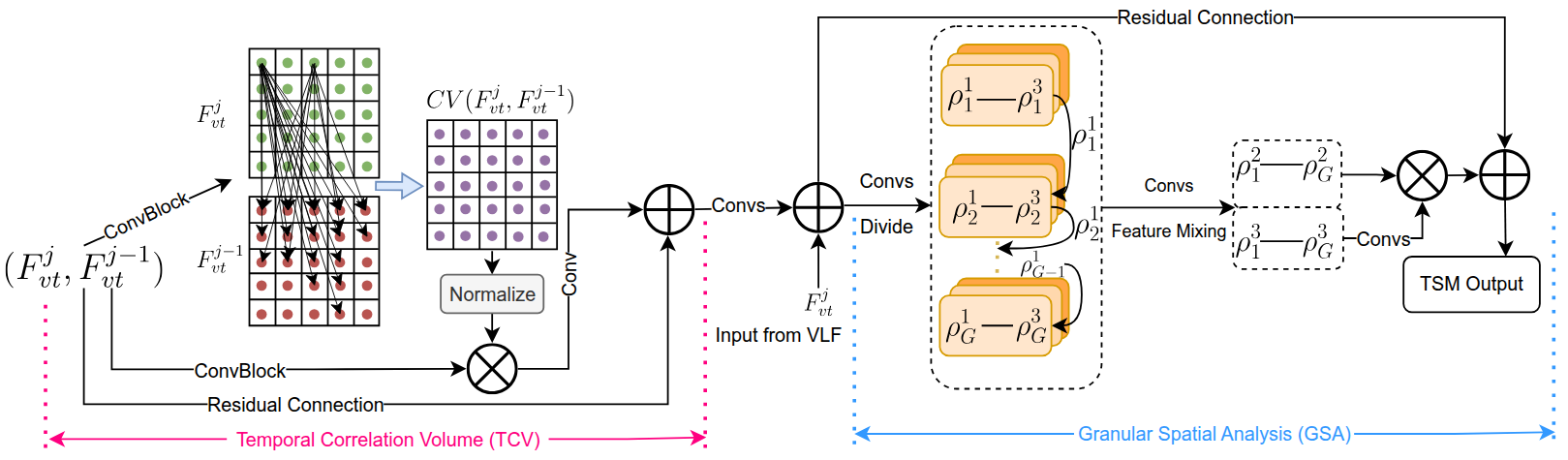}
    \caption{Framework of Temporal-Spatial Module (TSM), which is constituted by two parts, Temporal Correlation Volume (TCV) and Granular Spatial Analysis (GSA). The former is to find the target object through inter-frame changing, while the latter refines the intra-frame object features based on the TCV output.}
    \label{tsm_fig}
\end{figure*}

The resulting vision-language features, $F_{vt}$, are obtained by fusing the outputs of the vision and text encoders. After applying a linear projection to align feature dimensions, we fuse them through a dot product and then reshape the result to match the output shape of the vision encoder. The final fused feature, $F_{vt}$, is subsequently fed into the Temporal-Spatial Module (TSM) to capture motion and spatial information.

\subsection{Temporal-Spatial Module}
The Temporal-Spatial Module (TSM) is proposed to capture implicit motion information of the target object and refine it using the text-enhanced features produced by the Vision-Language Fusion (VLF) module. The architecture of TSM is illustrated in Fig. \ref{tsm_fig}. It consists of two key components: the Temporal Correlation Volume (TCV) and Granular Spatial Analysis (GSA).

The TCV is to obtain strong motion features through the temporal correspondence, based on the combination of vision-text features in frames $j$ and $j-1$, $(F_{vt}^j, F_{vt}^{j-1}) \in R^{2C \times H^{\prime} \times W^{\prime}}$, where $C$ is the channel number, $H^{\prime}$ and $W^{\prime}$ are height and width of the feature map. Inspired by \cite{cheng2022sltnet}, the TCV computes the correlation volumes between two feature maps of consecutive frames $F_{vt}^j$ and $F_{vt}^{j-1}$ to find subtle motion cues. The 4D correlation volume $CV(F_{vt}^j,F_{vt}^{j-1}) \in R^{H^{\prime} \times W^{\prime} \times H^\prime \times W^\prime}$, is as follows:
\begin{align}
CV(F_{vt}^j, F_{vt}^{j-1})_{xyuv} = \exp \left(\sum\limits_c {(F_{vt}^j)}_{xyc} \cdot {(F_{vt}^{j-1})}_{uvc}\right)
\end{align}
where $c$ denotes the channel index, and $xyuv$ are the coordinates of the current and adjacent frames. The $CV{(F_{vt}^j, F_{vt}^{j-1})}_{xyuv}$ denotes the correlation volumes between two elements at positions $(x,y)$ and $(u,v)$ in feature maps $F_{vt}^j$ and $F_{vt}^{j-1}$. Inconsistencies in the same positions $(x,y)$ between two feature maps indicate motion objects nearby. The Temporal Correlation Volume (TCV) module subsequently determines whether the motion in this area is relevant to the target object to distinguish and track. For each feature map $F_{vt}^j$ of a frame $j$, the TCV computes two correlation volumes between current frame and two adjacent frames, $CV(F_{vt}^j, F_{vt}^{j-1})$ and $CV(F_{vt}^j, F_{vt}^{j+1})$, to strengthen the reliability of the extracted features.

Normalization is then applied to the last two dimensions of $CV{(f_{t}, f_{t-1})}_{xyuv}$, $(u,v)$, to refine the relevant motion features between reference frames and the current frame. The normalization is formulated as follows:
\begin{align}
\widetilde{CV}(F_{vt}^j, F_{vt}^{j-1})_{xyuv} = \frac{CV(F_{vt}^j, F_{vt}^{j-1})_{xyuv}} {\sum_{u}\sum_{v} CV(F_{vt}^j, F_{vt}^{j-1})_{xyuv}}
\end{align}

Next, the normalized correlation volumes are aggregated with the channel-wise feature map $\lambda(F_{vt}^j, F_{vt}^{j-1}) \in R^{C \times H^{\prime} \times W^{\prime}}$, where $\lambda(\cdot)$ is a convolution layer that converts the channel number of $(F_{vt}^j, F_{vt}^{j-1})$, $2C$ to $C$. At last, the TCV obtains the final correlation volume $\varphi(F_{vt}^{j-1,j})$. The calculation of another correlation volume $\varphi(F_{vt}^{j+1,j})$ is the same as $\varphi(F_{vt}^{j,j-1})$. The aggregation of the channel-wise features and normalized correlation volumes is as follows:
\begin{align}
\varphi(F_{vt}^{j,j-1}) = \lambda(F_{vt}^j, F_{vt}^{j-1}) \cdot \widetilde{CV}(F_{vt}^j, F_{vt}^{j-1})_{xyuv} 
\end{align}

While the TCV captures inter-frame motion correlations, its output mainly indicates where feature changes occur between adjacent frames. These motion responses are useful for locating potential leak regions, but they are still insufficient for detailed intra-frame segmentation, especially because gas leaks are blurry, non-rigid, and weakly contrasted. Therefore, we introduce the Granular Spatial Analysis (GSA) module to further transform correlation-based motion cues into discriminative spatial features.

 The input of GSA $\tau(F_{vt}^{j,j-1})$ combines ($\varphi(F_{vt}^{j,j-1})$ and $F_{vt}^j$ from VLF as follows:
\begin{align}
\tau(F_{vt}^{j,j-1}) = \varphi(F_{vt}^{j,j-1}) + F_{vt}^j
\end{align}

This residual combination preserves the current-frame visual semantics while injecting temporal motion information. Then, a $1 \times 1$ convolution expands the channel dimension of $\tau(F_{vt}^{j,j-1})$, and the expanded $\tau(F_{vt}^{j,j-1})$ are divided into $g$ channel groups ($G_1-G_g$). Each group includes three channels of features $\rho_i^1, \rho_i^2, \rho_i^3$. Next, we split the first channel $\rho_i^1$ of $G_i$ to concatenate it to the last two layers of the next group of features ($\rho_{i+1}^2, \rho_{i+1}^3$). The overall processes are as follows:
\begin{equation}
 \begin{aligned}
     \rho_i^1 &= Split(Expand(G_i)) \\
      G_{i+1} &= Concat(G_{i+1}, \rho_i^1) \\
     \rho_{i+1}^1 &= Split(Expand(G_{i+1})) \\
 \end{aligned}
\end{equation}

GSA progressively transfers part of the feature from one group to the next, enabling interaction among different channel subspaces. This process functions as a lightweight multi-path refinement structure, allowing weak and fragmented leak-related responses to be gradually aggregated while suppressing isolated noise.

Functionally, this process works as a lightweight multi-path pyramid structure. The earlier groups preserve fine-grained local responses, while the later groups receive progressively enriched information from previous groups. This design is suitable for gas leak segmentation because leak regions often contain weak and fragmented responses distributed across different channels. By mixing channel groups progressively, GSA strengthens consistent leak-related patterns and suppresses isolated noise responses.

Finally, the second and third feature streams from all groups are aggregated into two feature sets, $N_2$ and $N_3$, through concatenation and convolution. The two sets of group-mixed features ($N_2$ and $N_3$) are fused through a convolutional block with a residual connection. The output of the GSA is denoted as $gsa_{vt}^j$. 
\begin{equation}
    \begin{aligned}
       N &= N_2\cdot N_3 \\
       gsa_{vt}^j &= \tau(F_{vt}^{j,j-1}) + Conv(N) \\
    \end{aligned}
\end{equation}

In a word, GSA bridges temporal correlation and dense mask decoding by refining coarse motion cues into granular spatial representations, improving the segmentation of blurry and deformable gas leak regions.

The TSM module effectively extracts valuable temporal and spatial cues across video frames and feature channels, enabling rich information interaction and progressive feature refinement. This module strengthens the model’s ability to perceive the target objects and provides a solid foundation for the final decoder predictions.

\subsection{Decoder and Post-Processing}
The decoder is a modified FPN that integrates multi-scale information by upsampling. Leveraging the rich features produced by the preceding modules, the FPN alone is sufficient to generate accurate segmentation masks. In contrast, using a separate COD head, like \cite{fan2021sinetv2}, may degrade the stability of the features. The effectiveness of our decoder design is further demonstrated in Section $4.6$.

Unlike typical VOS studies, which assume that target objects are present in every frame of a video \cite{xmempp,dshmp}, gas leaks do not appear sporadically throughout a video. Thus, the expected output of those non-leak frames should be completely black masks. However, significant noise results in false-positive masks for non-leak frames and makes many models fail in producing clean, completely black masks. To address the problem, we employ morphological opening \cite{soille2004opening} as a post-processing. Opening is defined as the dilation of the erosion of a binary mask by a kernel. The effectiveness of the opening depends on the kernel size. An oversized kernel may remove true-positive regions during the erosion stage, which cannot be restored during dilation, while an undersized kernel may fail to eliminate false positives. In our framework, the kernel size is adjustable according to different scenarios. Overall, this post-processing effectively suppresses false positives while preserving the integrity of true leak regions. 

\subsection{Loss Function}
To facilitate comparison with benchmark models, we adopt the commonly used loss function in the field of video object segmentation, Binary Cross-Entropy (BCE) and Intersection over Union (IoU). The BCE emphasizes pixel-level accuracy while the IoU captures global structure similarity. However, both loss functions treat all pixels equally and do not account for the varying importance of different regions. A more effective approach is to assign pixel-wise weights to specific areas, such as object boundaries and centers, which are typically more challenging to segment. If the value of pixel $(i,j)$ is different from its surrounding pixels, it indicates the place is more important. Thus, the place in this frame will be largely weighted. The formulas for weighted BCE $L_{bce}^w$ and weighted IoU $L_{iou}^w$ are the following:
\begin{equation}
\begin{aligned}
L_{\mathrm{bce}}^w &= \frac{\sum_{k=1}^{HW} (1+\alpha w_k)\,\big[y_k\log p_k + (1-y_k)\log(1-p_k)\big]}
         {\sum_{k=1}^{HW} (1+\alpha w_k)} \\
L_{\mathrm{iou}}^w &= 1-\frac{\sum_{k=1}^{HW} (1+\alpha w_k)\,(y_k p_k)}
            {\sum_{k=1}^{HW} (1+\alpha w_k)\,\big(y_k+p_k-y_k p_k\big)} .
\end{aligned}
\end{equation}
where $\alpha$ is the hyperparameter, $p_{ij}$ and $gt_{ij}$ are the prediction probability and the ground truth pixel at coordinate $(i, j)$. $w_{ij}$ is the weight for each pixel. Inspired by \cite{wei2020f3net}, we add weights to each position of a ground truth frame in both loss functions as follows:
\begin{align}
w_{ij} = \left| \frac{\sum_{S_{ij}} gt}{\sum_{S_{ij}}1}-gt_{ij} \right|
\end{align}
where $S_{ij}$ is the surrounding area of the pixel $(i,j)$. In a word, weighted BCE is assigned weights for pixels to focus on critical areas. The weighted IoU loss is a crucial component for video object segmentation. It highlights challenging pixels while retaining attention to the overall structural accuracy of the object masks. Finally, our loss function combines the wBCE and wIoU, defined as follows:
\begin{align}
L = L_{bce}^{w} + L_{iou}^{w}
\end{align}

\begin{table*}[t]
    \centering
    \caption{Quantitative comparison with benchmark methods on two datasets.}
    \captionsetup{skip=5pt}
    \scalebox{1.3}{
    \begin{tabular}{lcc|ccc|ccc}
    \toprule
    \multirow{2}{*}{Models} & \multirow{2}{*}{Input Type} & \multirow{2}{*}{Year} & \multicolumn{3}{c}{SimGas (Avg)} & \multicolumn{3}{c}{IGS-Few} \\
    \cmidrule(lr){4-6}  \cmidrule(lr){7-9} 
    & &  &$J\uparrow$ &$F\uparrow$ &$J\&F\uparrow$ &$J\uparrow$ &$F\uparrow$ &$J\&F\uparrow$ \\
    \midrule
    SINet-V2 \cite{fan2021sinetv2}    &Image &2021 &37.29 &42.76 &40.03 &17.45 &28.15 &22.80 \\
    GasSeg \cite{yu2024gasseg}        &Image &2024 &37.35 &46.16 &41.76 &55.40 &67.00 &61.20 \\
    \midrule
    SLT-Net \cite{cheng2022sltnet}    &Video &2022 &39.24 &44.40 &41.82 &66.43 &76.52 &71.47 \\
    Cutie \cite{cheng2024cutie}       &Video &2024 & -    & -    & -    &46.61 &59.07 &52.84 \\
    Zoomnext \cite{pang2024zoomnext}  &Video &2024 &21.67 &27.10 &24.39 &59.28 &70.43 &64.86 \\
    LangGas \cite{guo2025langgas}     &Video &2025 &53.57 &63.95 &58.76 &10.66 &15.16 &12.91 \\
    FGSTP \cite{zhao_fgstp}                &Video &2025 &39.45 &44.42 &41.94 &66.70 &76.56 &71.63 \\
    \textbf{JVLGS (Ours)}             &Video &2025 &\textbf{61.73} &\textbf{70.25} &\textbf{65.99} &\textbf{67.06} &\textbf{77.05} &\textbf{72.05} \\
    \bottomrule
    \end{tabular}
    }
    \label{expres}
\end{table*} 

\section{Experiments}
\subsection{Datasets}
We evaluate our model on two datasets. The first, SimGas, is a synthetic dataset. Its backgrounds are partially sourced from a real-world dataset, GasVid \cite{wang2022videogasnet}, while the rest are generated using DALL·E 2 \cite{dalle2}. Foreground gas leaks are simulated using Blender. Manual annotation of GasVid frames is challenging due to the difficulty of precisely delineating leak regions, and the limited variety of leak scenarios constrains the evaluation of model generalizability. In comparison, the SimGas dataset provides more diverse leak scenarios that better represent real-world conditions. Notably, it preserves precise annotations by compositing realistically rendered leaks and occluding foreground elements onto backgrounds. The dataset comprises 31 videos, totaling approximately 1,600 frames. Additional details are provided in \cite{guo2025langgas}.

The second dataset is derived from the IGS dataset \cite{yu2024gasseg}, which contains 4,453 video frames for training and 1,897 for testing across a total of 27 videos. In the original dataset split, training and testing frames are drawn from the same videos, with the majority of frames from each video used for training and the remaining frames reserved for testing. However, video surveillance systems always adopt few-shot training, which uses only a few frames per video for training, while the majority are reserved for testing. To better reflect real-world leak monitoring conditions, we adopt this few-shot training strategy. Specifically, we cap the number of training frames per video, leaving hundreds of frames per video for testing. We refer to this reconfigured dataset as IGS-Few.

\subsection{Implementation Details}
For fair comparisons across methods, all experiments are conducted using the same software and hardware configurations. We set the experimental image resolution to 352$\times$352. We train for 60 epochs on SimGas and 150 epochs on the IGS-Few dataset. Moreover, we set the initial learning rate (lr) to $10^{-4}$, which decays to $10^{-5}$ in the final 20\% of training epochs. For the loss function hyperparameter, we set the $\alpha$ to 0.5 in all experiments. Following standard practices in VOS evaluation, we adopt metrics that assess both region similarity and boundary accuracy between the predicted masks and the ground truth annotations. Empirically, we adopt the Jaccard Index ($J$), contour accuracy ($F$), and their average ($J\&F$), following \cite{vosbenchmark}, to evaluate mask overlap and boundary alignment. All experiments are performed on an NVIDIA RTX 4090 GPU with 24 GB of memory.

\subsection{Results}
\subsubsection{Experimental Setup}
When comparing our model with other SOTA methods, we observe significant variations in both the evaluation metrics and dataset splits used by different models. LangGas \cite{guo2025langgas} computes $J$ and $F$ for each video by confusion matrices, while others calculate the average $J$ and $F$ scores in all frames of the test set. GasSeg \cite{yu2024gasseg} uses most frames for training and minors for testing, and it also averages background and foreground results in the metric evaluation. To ensure a fair comparison, we standardize the metric computation for all benchmark models in $4.3.2$. Additionally, we report results using each method’s original metric computation or dataset split when conducting individual comparisons in $4.3.3$.

\subsubsection{Comparison with SOTA Methods}
\begin{figure*}[t]
    \centering
    \includegraphics[width=1\textwidth]{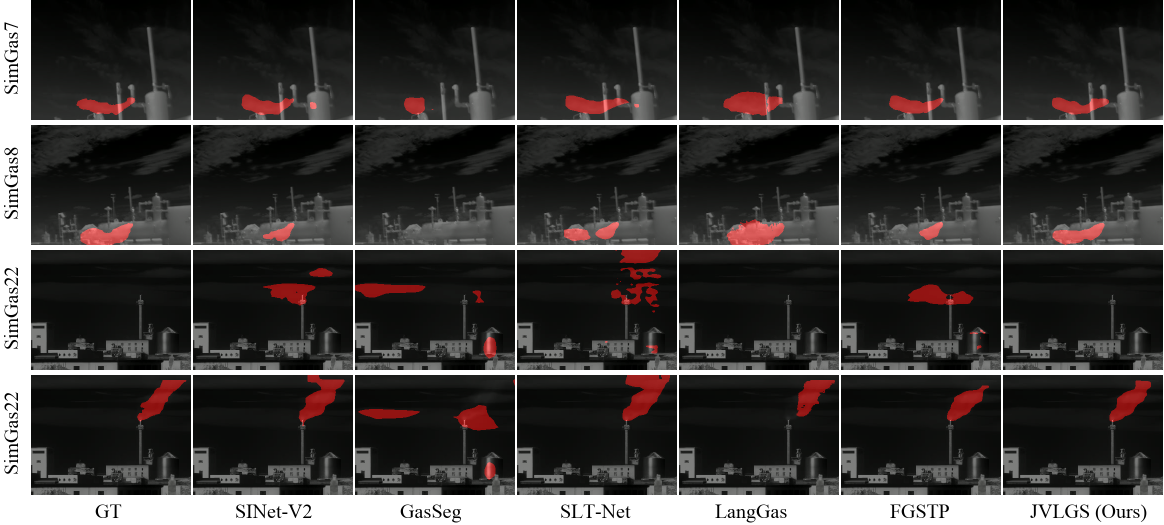}
    \caption{Visualized results on SimGas. Our model achieves the highest accuracy and generates whole-black masks for non-leak cases.}
    \label{simgas_vis}
    \vspace{10pt}
    \includegraphics[width=1\textwidth]{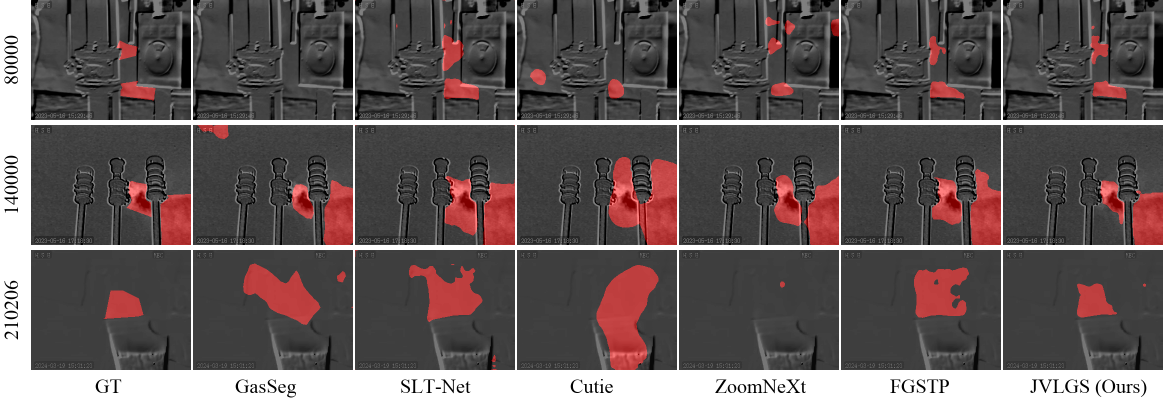}
    \caption{Visualized results on IGS-Few. Our model distinguishes the leak better than other models in few-shot learning.}
    \label{igsfew_vis}
\end{figure*}

Table \ref{expres} summarizes the results using our unified metric computation. And Fig. \ref{simgas_vis} and Fig. \ref{igsfew_vis} present qualitative visualizations on the SimGas and IGS-Few datasets. SINet-V2 is an early method for camouflaged object detection and does not have a temporal module. Consequently, its performance is lower than many other methods on both datasets. GasSeg effectively finds leaks by fusing local and global information. It performs well on its proposed dataset, but it struggles on SimGas, resulting in a low $J\&F$ of 41.76. SLT-Net and FGSTP demonstrate comparable performance on both the SimGas and IGS-Few datasets. However, they suffer from false positives in non-leak frames, often generating random artifacts. ZoomNeXt focuses on capturing multi-scale spatial features but overlooks motion information. Thus, it underperforms on the SimGas videos. Cutie depends on the mask of the first frame to initialize the target object and guide segmentation in subsequent frames. The absence of leaks in the initial frames poses a huge obstruction to the SimGas dataset prediction. LangGas combines background subtraction with a large language model for mask prediction. But it struggles with the dithering backgrounds. Our proposed method, JVLGS, integrates motion, spatial, and textual prompts to enhance feature representation. It achieves a $J\&F$ score of 65.99 on SimGas and 72.05 on IGS-Few, outperforming all other methods on both datasets. 


\begin{table}[hbpt]
    \centering
    \captionsetup{skip=5pt}
    \setlength{\tabcolsep}{10pt}
    \caption{Comparison with LangGas using its original metric calculation, on the SimGas dataset.}
    \begin{tabular}{c|ccc}
    \toprule
    Models &$J\uparrow$ &$F\uparrow$ &$J\&F\uparrow$\\
    \midrule
    LangGas \cite{guo2025langgas}   &58.82 &72.82 &65.82 \\
    \textbf{JVLGS (Ours)}           &\textbf{69.54} &\textbf{79.75} &\textbf{74.65}\\
    \bottomrule
    \end{tabular}
    \label{originalalgorithm}
\vspace{15pt}
    \centering
    \captionsetup{skip=5pt}
    \caption{Comparison with GasSeg using its original split, on the IGS dataset.}
    \begin{tabular}{c|ccc}
    \toprule
    Models &$J\uparrow$ &$F\uparrow$ &$J\&F\uparrow$\\
    \midrule
    GasSeg \cite{yu2024gasseg}    &78.21 &86.42 &82.31 \\
    \textbf{JVLGS (Ours)}         &\textbf{81.82} &\textbf{89.01} &\textbf{85.42}\\
    \bottomrule
    \end{tabular}
    \label{originalsplit}
\end{table}

\subsubsection{Different Metrics Comparison with Some Models}
\vspace{-5pt}
As noted before, LangGas and GasSeg use different evaluation metrics and dataset splits from ours. To ensure fairness, we re-implement their calculations to evaluate our method and LangGas. Table \ref{originalalgorithm} presents results based on LangGas’s metric computation, while Table \ref{originalsplit} shows performance under GasSeg’s dataset split. In both settings, our method consistently outperforms the others, demonstrating its robustness across evaluation protocols.

\subsection{Analysis of Text Prompt Effect}
\begin{figure}[hbpt]
    \includegraphics[width=1\linewidth]{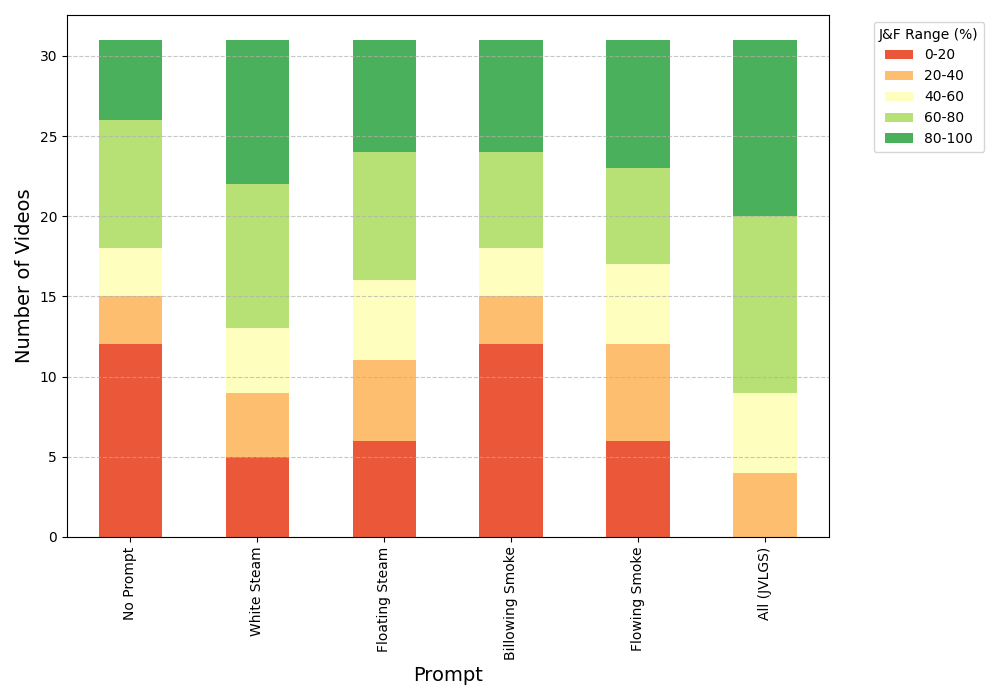}
    \caption{$J\&F$ value distributions of each prompt on SimGas dataset videos.}
    \label{simgas_bar}
\end{figure}

\begin{figure}
    \centering
    \includegraphics[width=1\linewidth]{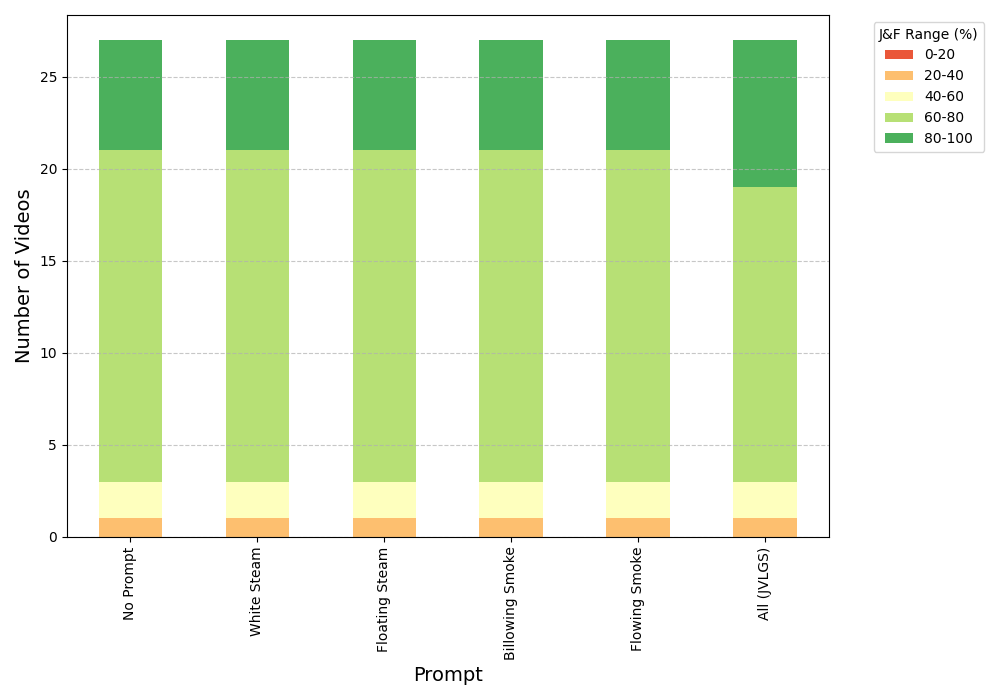}
    \caption{$J\&F$ value distributions of each prompt on IGS-Few dataset videos.}
    \label{igsfew_bar}
\end{figure}

We further verify the effectiveness of the language prompt. Specifically, we calculate the improvements of $J\&F$ by different prompts for each video on both datasets. The reason we select these prompts is illustrated in Section $3.2$. Figures \ref{simgas_bar} and \ref{igsfew_bar} illustrate the performance distribution across all prompts for both datasets. The vertical axes represent the number of videos, each displayed in a different color corresponding to a specific range of $J\&F$ values. We categorize the $J\&F$ values into five groups: red for 0–20 (lowest), orange for 20–40 (poor), yellow for 40–60 (moderate), light green for 60–80 (suboptimal), and green for 80–100 (best).

\begin{figure*}[hbpt]
    \centering
    \includegraphics[width=1.0\linewidth]{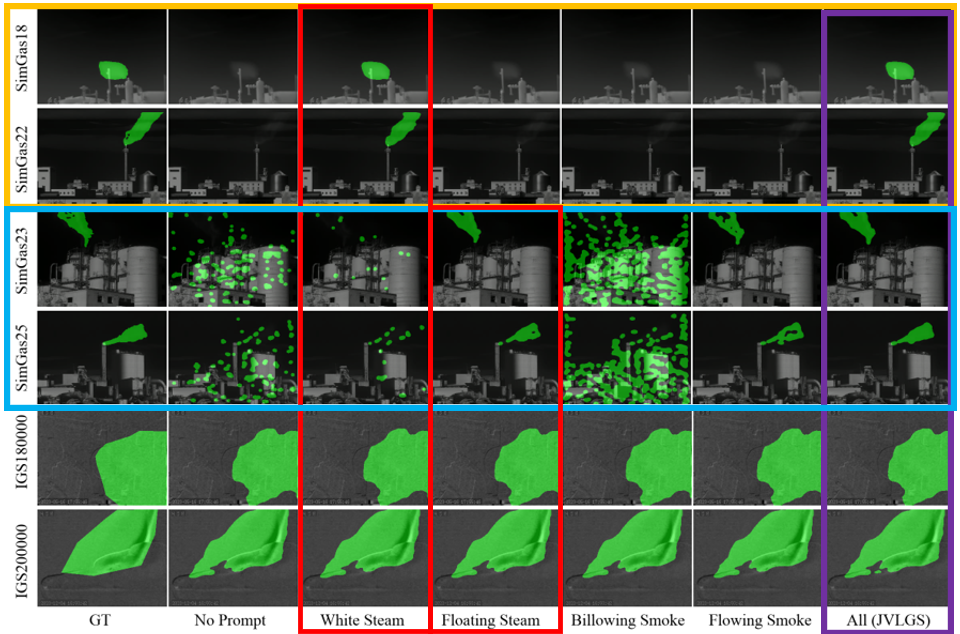}
    \caption{Visualized contribution of text prompt. Each prompt word tends to enhance performance in some specific scenarios. e.g, “White Steam” contributes significantly to videos 18 and 22. “Floating Steam” contributes to videos 23 and 25. }
    \label{vis_prompte_con}
\end{figure*}

Figure \ref{simgas_bar} illustrates the effect of different text prompts on segmentation performance. On the SimGas dataset, the prompt “White Steam” provides the most substantial improvement, markedly reducing the number of videos with $J\&F$ scores in the 0-20 range and shifting them toward higher-performance intervals. This suggests that semantically precise prompts can guide the model to focus on more discriminative visual cues, such as the color, texture, and semi-transparent appearance of gas-like targets, thereby improving feature representation for difficult samples. Prompts such as “Floating Steam” and “Flowing Smoke” also contribute notably, indicating that motion-related textual descriptions help the model capture dynamic patterns and boundary variations of vapor-like objects. In contrast, “Billowing Smoke” mainly improves easier-to-segment videos, as reflected by the longer deep-green segment compared with the no-prompt baseline, which implies that this prompt may strengthen high-level semantic activation but is less effective for ambiguous or low-contrast cases.

The influence of text prompts is much weaker on the IGS-Few dataset, as shown in Figure \ref{igsfew_bar}. No single prompt leads to a clear performance gain, while the combination of prompts only brings a slight improvement. One possible reason is that, under the few-shot setting, the model has already observed similar target appearances during training and can learn task-specific visual patterns from the annotated support samples. As a result, external textual guidance becomes less critical, and the marginal benefit of prompts is reduced. In addition, when visual supervision is sufficiently informative, text prompts may provide overlapping semantic information rather than introducing new discriminative cues.

Table \ref{prompt_contribution} quantifies the contribution of each prompt on the SimGas dataset. We define prompt contribution as the average increase in $J\&F$ score relative to the non-prompt baseline, calculated across all videos. The column “Improved Videos” indicates the number of videos significantly influenced by each prompt. A video is considered “improved” if the improvement in its $J\&F$ score exceeds 0.5 points.

Moreover, Figure \ref{vis_prompte_con} further visualizes the impact of each prompt. The video frames located at the intersection of red, yellow, and blue rims indicate that different prompts benefit specific subsets of videos. Each prompt activates particular semantic and visual attributes, such as color, shape, motion, or diffusion patterns, and these attributes are useful only when they match the target characteristics of a given video. Therefore, relying on a single prompt may bias the model toward a limited feature space, whereas using multiple prompts encourages the model to learn more comprehensive and robust multimodal representations. Ultimately, all four prompt words collectively enhance feature representations on the SimGas dataset, ensuring that each gas leak video is guided by at least one effective prompt. However, the last two rows of the figure indicate that the prompts have only a slight influence on predictions for the IGS-Few dataset. Thus, the language prompts are primarily beneficial for unseen scenarios. In contrast, the proposed model achieves reliable segmentation in seen scenarios even under few-shot training without the need for additional language prompts.

\begin{table}[tbph]
    \centering
    \caption{Contribution of each prompt for JVLGS performance on the SimGas dataset.}
    \scalebox{0.95}{
    \begin{tabular}{ccccccc}
    \toprule  
      Prompts & Avg. $J\&F$ Contribution &No. of Improved Videos \\
    \midrule
    No Prompt            &0     &0  \\
    “White Steam”        &15.03 &21 \\
    “Floating Steam”     &10.26 &12 \\
    “Billowing Smoke”    &1.14  &12 \\
    “Blowing Smoke”      &8.17  &18 \\
    All (JVLGS)          &26.35 &28 \\
    \bottomrule
    \end{tabular}
    }
    \label{prompt_contribution}
\end{table}

\vspace{-10pt}
\subsection{Ablation Studies}
Table \ref{ablation} presents the results of our ablation studies conducted on both the SimGas and IGS-Few datasets. We evaluate the individual contributions of the VLF module, FPN decoder, and post-processing.
\vspace{5pt}

\textbf{Effectiveness of Text Prompts.} We assess the impact of the language prompt module while keeping all other components unchanged. As shown in Rows 4 and 6 of Table \ref{ablation}, removing the VLF module results in performance degradation on both datasets, with a more noticeable drop on SimGas and a slight decrease on IGS-Few. This observation highlights the greater contribution of the VLF module on the SimGas dataset compared to IGS-Few.
\vspace{5pt}

\begin{table*}[ht]
    \centering
    \caption{Ablation studies on both datasets demonstrating the effectiveness of each component within our proposed model.}
    \scalebox{1.3}{
    \begin{tabular}{cccc|ccc|ccc}
    \toprule
    \multirow{2}{*}{TSM} & \multirow{2}{*}{VLF} & \multirow{2}{*}{Decoder} & \multirow{2}{*}{Post-Processing}& \multicolumn{3}{c}{SimGas (Avg)} & \multicolumn{3}{c}{IGS-Few} \\
    \cmidrule(lr){5-7}  \cmidrule(lr){8-10} 
    & & &  &$J\uparrow$ &$F\uparrow$ &$J\&F\uparrow$ &$J\uparrow$ &$F\uparrow$ &$J\&F\uparrow$ \\
    \midrule
    \checkmark &           &NCD &             &39.45 &44.42 &41.94 &66.70 &76.56 &71.63  \\
    \checkmark &\checkmark &NCD &             &38.88 &44.24 &41.55 &66.90 &76.96 &71.93  \\
    \checkmark &\checkmark &NCD &\checkmark   &57.73 &66.05 &61.90 &66.94 &76.97 &71.95  \\
    \checkmark &           &FPN &\checkmark   &59.21 &67.26 &63.23 &66.97 &\textbf{77.08} &72.03  \\
    \checkmark &\checkmark &FPN &             &39.06 &44.50 &41.78 &67.02 &77.05 &72.03  \\
    \checkmark &\checkmark &FPN &\checkmark   &\textbf{61.73} &\textbf{70.25} &\textbf{65.99} &\textbf{67.06} &77.05 &\textbf{72.05}  \\
    \bottomrule
    \end{tabular}
    }
    \label{ablation}
\end{table*}

\begin{figure*}[hbpt]
    \includegraphics[width=1.0\linewidth]{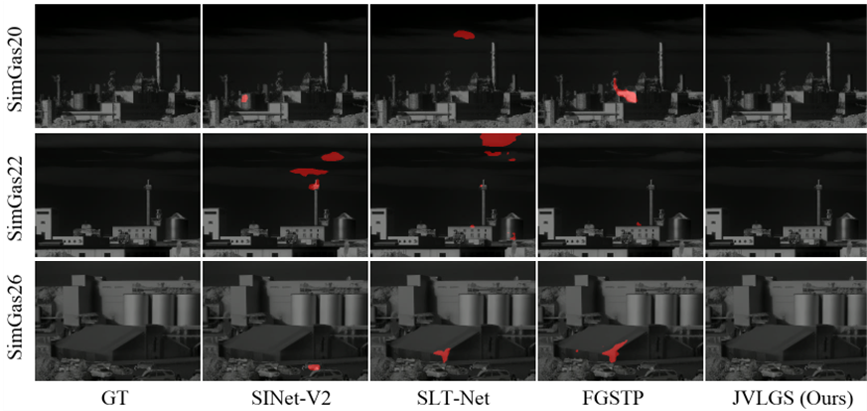}
    \caption{False-positive removal results. Our model suppresses false positives on non-leak frames and produces masks only when leaks are present.}
    \label{solved_vljgs}
\end{figure*}

\textbf{Decoder Choice.} The differences between Rows 2 and 5, as well as Rows 3 and 6 in Table \ref{ablation}, underscore the importance of decoder choice. Under identical conditions, regardless of the presence of the VLF module and post-processing, models using the FPN decoder consistently outperform those using the NCD decoder \cite{zhao_fgstp} on both datasets. This can be attributed to FPN’s straightforward multi-scale feature integration. In contrast, the NCD involves more complex structures, such as reversing and inserting feature maps, which may lead to incompatibilities with other modules and ultimately hinder overall performance.
\vspace{5pt}

\textbf{Effectiveness of Post-Processing.} The ablation studies also quantify the improvement by post-processing. Comparison in Table \ref{ablation} between rows 2 \& 3, 5 \&6 shows that the elimination of false positives on non-leak frames significantly improves the segmentation accuracy. Fig. \ref{solved_vljgs} visualizes the effectiveness of our proposed model in eliminating false positives. Previous methods (Columns 2–4) frequently produce false positives on non-leak frames. This often occurs when clouds are mistakenly identified as leaks, as seen in videos SimGas 20 \& 22, due to their visual similarity. Even in non-cloud scenes, such as video SimGas 15, these methods incorrectly highlight the fluctuating luminance as objects. However, the results of our model do not have false positives on non-leak frames, and the stability of model predictions has been greatly improved.
\vspace{5pt}

\begin{figure}[hbpt]
    \includegraphics[width=1.0\linewidth]{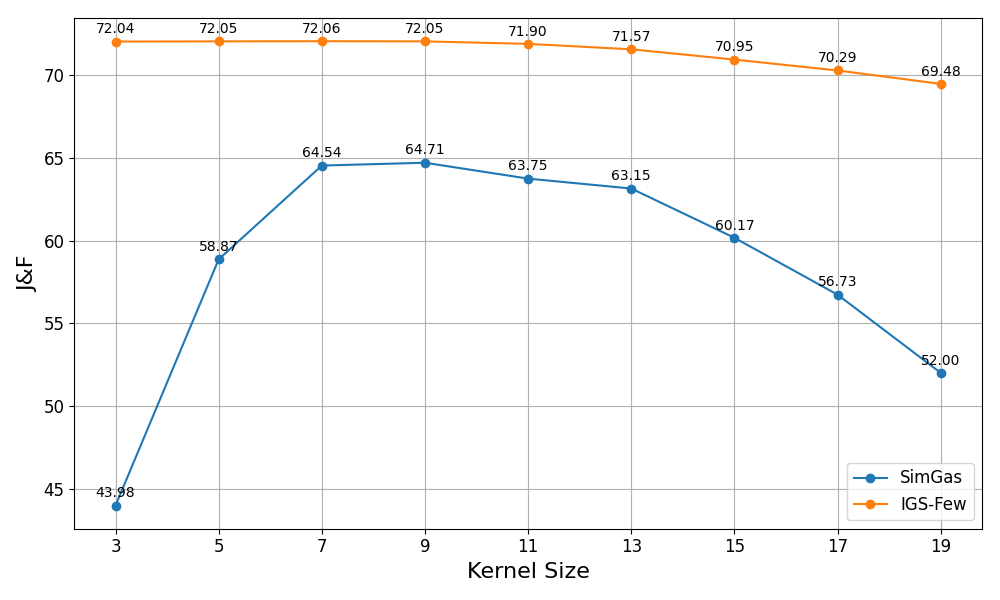}
    \caption{The relationship between $J\&F$ score and kernel size in our model.}
    \label{opening}
\end{figure}

\textbf{Coordination Between Modules.} An interesting observation from Table \ref{ablation} is that Rows 1 and 2 appear to suggest a negative impact of the VLF module on the SimGas dataset. Even when the decoder is switched to FPN in Row 5, the performance remains lower than that of Row 1, which uses the NCD decoder without VLF. This discrepancy can be attributed to interactions among components: the VLF module performs better when paired with the FPN decoder, as evidenced by the improvements between Rows 2 and 5, Rows 3 and 6. Notably, the combination of VLF, FPN, and post-processing in Row 6 achieves the highest scores on both datasets, underscoring the importance of effective coordination between modules.

\subsection{Post-Processing Optimal Kernel Size} 
We also explore the optimal kernel size for the post-processing, with results shown in Figure \ref{opening}. The blue line is $J\&F$ on the SimGas dataset, and the orange one is on the IGS-Few dataset. Both excessively small and large kernels degrade performance: small kernels fail to remove false positives, while large kernels tend to over-erode valid mask regions, making them unrecoverable during dilation. Our experiments identify a kernel size of 9 as delivering the best performance across both datasets, and we adopt this value as a hyperparameter in our model.

\subsection{Performance-Efficiency Balance} 
\begin{figure}[hbpt]
    \centering
    \includegraphics[width=0.95\linewidth]{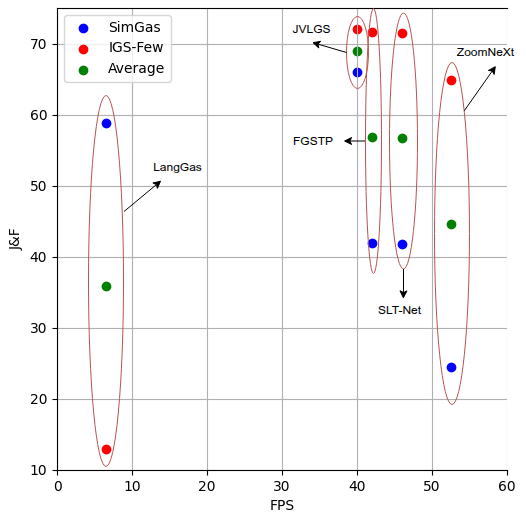}
    \caption{Performance-Efficiency Diagram. Blue and red points represent results on our two datasets, SimGas and IGS-Few, while green points are the average accuracy across both datasets.}
    \label{perate}
\end{figure}

Finally, we analyze the computational efficiency and model complexity of the proposed method. As shown in Fig. \ref{perate}, JVLGS achieves a favorable trade-off between segmentation accuracy and processing speed. Among all benchmark models, our method attains superior performance, achieving approximately 70 $J\&F$ while sustaining a real-time inference speed of 40 FPS. In addition, the proposed model contains approximately 145M parameters, providing an objective measure of its computational complexity. Despite this model scale, JVLGS maintains efficient inference. This balanced combination of high accuracy, real-time performance, and manageable model complexity makes JVLGS well-suited for practical industrial monitoring applications.

\section{Conclusion}
In this paper, we presented JVLGS, a novel framework designed to achieve high-accuracy gas leak detection and segmentation. By integrating the complementary strengths of visual features and text prompts, JVLGS enhances feature representation and significantly improves segmentation precision. Extensive experimental results demonstrate that JVLGS achieves superior mask quality under both supervised and few-shot learning settings, outperforming existing state-of-the-art methods. Furthermore, an adaptive post-processing module effectively suppresses false positives in non-leak scenarios, ensuring greater robustness and reliability in real-world industrial environments.

Overall, JVLGS offers a practical and efficient solution for industrial gas leak monitoring and safety assurance. In future work, we plan to extend JVLGS to additional video datasets and further explore its potential for real-time industrial surveillance in real world deployments.


\vspace{15pt}

\noindent\textbf{Author Contributions} Xinlong Zhao contributed to methodology development, model implementation, data curation, experiments, result analysis, and manuscript drafting. Qixiang Pang contributed to technical guidance and critical revision of the manuscript. Shan Du contributed to study conception, research guidance, result interpretation, and critical revision of the manuscript. All authors reviewed the manuscript, approved the final version, and agreed to be accountable for all aspects of the work.

\vspace{15pt}

\noindent\textbf{Data Availability} All data generated or analyzed during this study are included in this article.
\vspace{20pt}

\noindent\textbf{Acknowledgements} This work is supported by the Social Sciences and Humanities Research Council of Canada (SSHRC) under grant NFRF-GR024801.

%
\section*{Declaration}
\textbf{Conflict of interest} The authors declare that they have no conflict of interest.


\bibliographystyle{spmpsci}      
\bibliography{ref}   
\vspace{30pt}

\noindent
\begin{minipage}[t]{0.3\columnwidth}
    \vspace{0pt}
    \includegraphics[width=\linewidth]{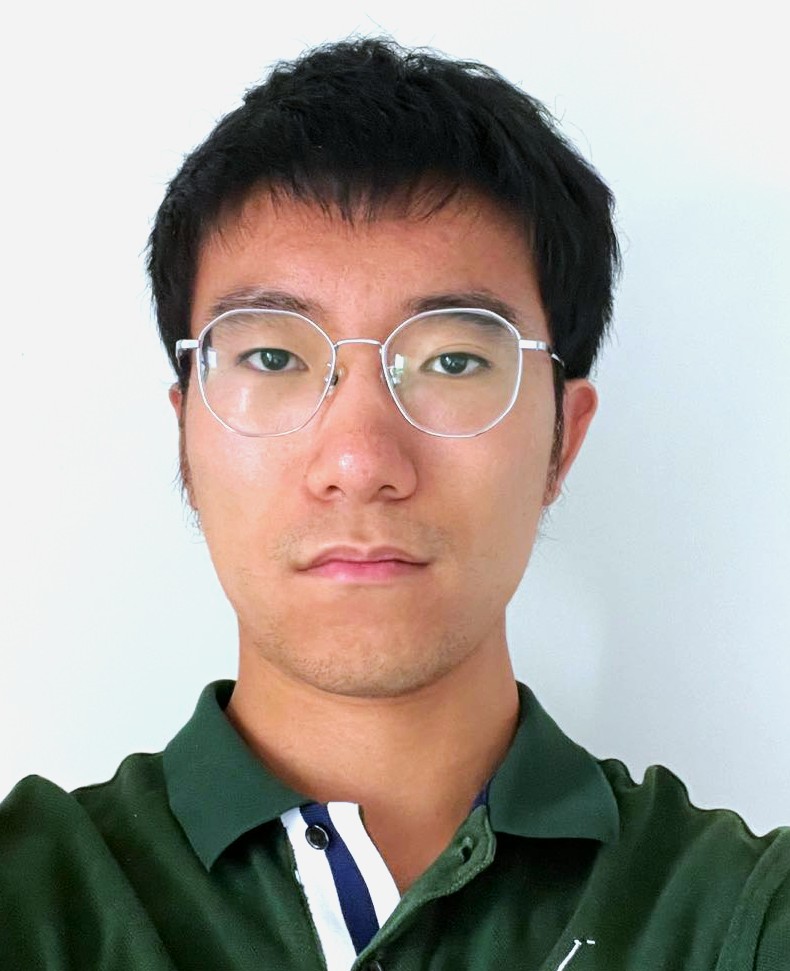}
\end{minipage}
\hfill
\begin{minipage}[t]{0.65\columnwidth}
    \vspace{0pt}
    \textbf{Xinlong Zhao} received his Master’s degree in Computer Science from the University of British Columbia in 2025. He is an IEEE Student Member. He is currently a Computer Vision Algorithm Engineer at Qihoo 360 in Beijing, China. His work focuses on developing advanced computer vision algorithms and deploying large-scale AI systems for real-world applications.
    
    His research interests include Vision–Language Models, video inpainting, and camouflaged video object detection. He has also worked on efficient AI inference architecture deployment. His broader interest lies in building practical AI systems that bridge academic research and cutting-edge industrial applications.
\end{minipage}

\vspace{15pt}

\noindent
\begin{minipage}[t]{0.3\columnwidth}
    \vspace{0pt}
    \includegraphics[width=\linewidth]{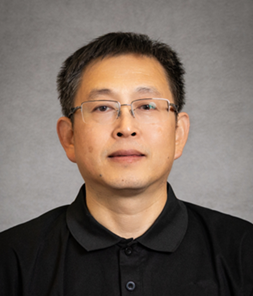}
\end{minipage}
\hfill
\begin{minipage}[t]{0.65\columnwidth}
    \vspace{0pt}
\textbf{Qixiang Pang} received his Ph.D. from Beijing University of Posts and Telecommunications and completed postdoctoral research in Electrical and Computer Engineering at the University of British Columbia. He is currently an Assistant Professor in the Department of Computer Science and Cybersecurity at the University of Central Missouri, USA. Previously, he worked at UBC, Lakehead University, and the Chinese University of Hong Kong, and in industry with Motorola, General Dynamics, and TELUS. His research focuses on computer vision, and cloud computing. He has served on technical program committees for over 50 international conferences and is an invited Editor for the Computers journal.  
\end{minipage}

\vspace{15pt}

\noindent
\begin{minipage}[t]{0.3\columnwidth}
    \vspace{0pt}
    \includegraphics[width=\linewidth]{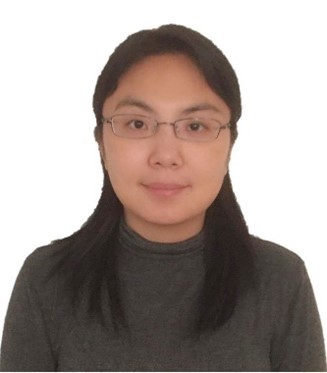}
\end{minipage}
\hfill
\begin{minipage}[t]{0.65\columnwidth}
    \vspace{0pt}
\textbf{Shan Du} is an assistant professor of Computer Science at the University of British Columbia, Okanagan. She received the Ph.D. degree in Electrical and Computer Engineering from UBC Vancouver in 2009. Before joining UBC, she worked as an assistant professor with Lakehead University and as a Research Software Engineer with IntelliView Technologies Inc. Shan has 15+ years research experience on computer vision, image and video processing, and machine learning. Shan serves as an Associate Editor of IEEE Trans. on Multimedia and IEEE Trans. on Circuits and Systems for Video Technology, Session Chair of ICIP 2019 – 2026, and TPC co-chair of GI 2025.
\end{minipage}

\end{document}